\documentclass[fleqn,10pt]{SelfArx} 

\usepackage[english]{babel} 

\graphicspath{{../final_version/source_files/}}

\newcommand{\ea}{et al. }
\newcommand{\eg}{e.g. }
\newcommand{\ie}{i.e. }

\definecolor{color1}{RGB}{0,0,90} 
\definecolor{color2}{RGB}{0,20,20} 

\usepackage{subfigure}
\usepackage{url}
\usepackage{graphicx}
\usepackage{multirow}

\JournalInfo{ $\quad$ } 
\Archive{ $\quad$ }

\PaperTitle{Predicting Human Intentions from Motion Only: \\ A 2D+3D Fusion Approach} 

\Authors{Andrea Zunino$^{\pmb{1},\pmb{3}}$, Jacopo Cavazza$^{\pmb{1},\pmb{3}}$, Atesh Koul$^{\pmb{2}}$, \\ Andrea Cavallo$^{\pmb{2}}$, \\ Cristina Becchio$^{\pmb{2}}$ and Vittorio Murino$^{ \pmb{1},\pmb{4}}$\\ 
	{ firstname.lastname@iit.it} \\
	$^{\pmb 1}$ Pattern Analysis \& Computer Vision (PAVIS) -- Istituto Italiano di Tecnologia -- \textit{Genova, Italy} \\
	$^{\pmb 2}$ C’MON Cognition, Motion and Neuroscience -- Istituto Italiano di Tecnologia (IIT) -- \textit{Genova, Italy} \\
	$^{\pmb 3}$ Electrical, Electronics and Telecommunication Engineering and Naval Architecture Department \\ (DITEN) -- Universit\`{a} degli Studi di Genova --  \textit{Genova, Italy}\\
	$^{\pmb 4}$Computer Science Department -- Universit\`{a} di Verona --  \textit{Verona, Italy}}

\newcommand{\keywordname}{Keywords} 

\Abstract{In this paper, we address the new problem of the prediction of human intents. There is neuro-psychological evidence that actions performed by humans are anticipated by peculiar motor acts which are discriminant of the type of action going to be performed afterwards. In other words, an actual intent can be forecast by looking at the kinematics of the immediately preceding movement. To prove it in a computational and quantitative manner, we devise a new experimental setup where, without using contextual information, we predict human intents all originating from the same motor act. We posit the problem as a classification task and we introduce a new multi-modal dataset consisting of a set of motion capture marker 3D data and 2D video sequences, where, by only analysing very similar movements in both training and test phases, we are able to predict the underlying intent, i.e., the future, never observed action. We also present an extensive experimental evaluation as a baseline, customizing state-of-the-art techniques for either 3D and 2D data analysis. 
Realizing that video processing methods lead to inferior performance but show complementary information with respect to 3D data sequences, we developed a 2D+3D fusion analysis where we achieve better classification accuracies, attesting the superiority of the multimodal approach for the context-free prediction of human intents.}

\begin{document}

\flushbottom 

\maketitle 

\tableofcontents 

\thispagestyle{empty} 


\section{Introduction}
Action and activity\footnote{If not differently specified, activity and action are here used interchangeably.} recognition are surely intriguing and most active areas in computer vision. The task here consists in the classification of \emph{fully observed} action or activity. More recently, the scientific community has also investigated a variant, extending the paradigm to the ``early'' activity recognition, that is, recognizing an action before it is fully disclosed. 
Early activity recognition is sometimes improperly confused with action prediction: this happens when, instead of predicting a future action, the latter is recognized by just classifying its very beginning. 
The actual action prediction problem consists instead in the classification of future actions considering all the events occurring up to a certain instant \cite{Amit:ACCV14}. As a different paradigm, here we aim at introducing a brand new action prediction challenge consisting in the prediction of human \emph{intents}, defined as the overarching goal embedded in an action sequence. 

\begin{figure*}[t!]
	\centering
	\subfigure[Action/activity recognition\label{s:AR}]{\includegraphics[height=0.12\textheight,width=.38\textwidth]{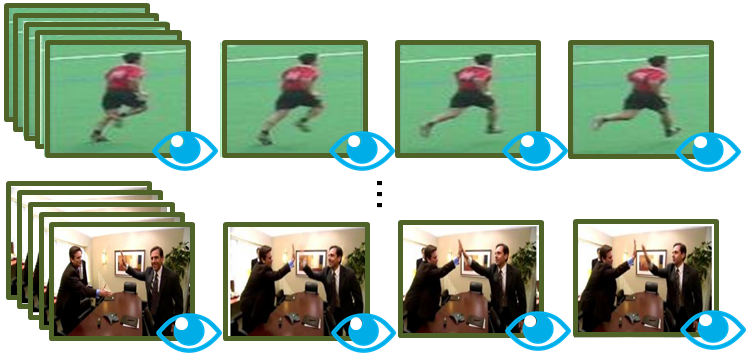}}\qquad\quad
	\subfigure[Early activity recognition\label{s:EAR}]{\includegraphics[height=0.12\textheight,width=.4\textwidth]{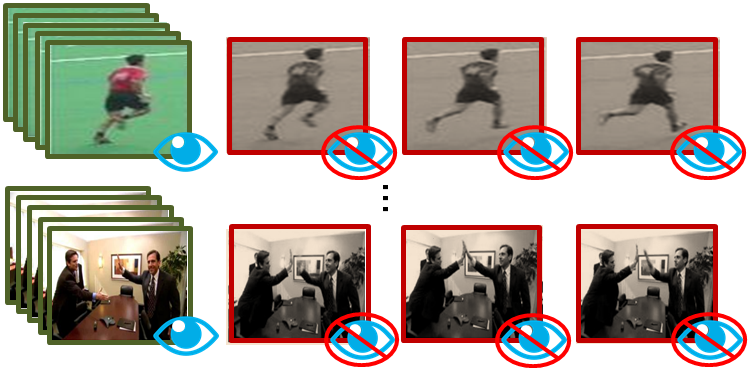}}\\
	\subfigure[Action prediction\label{s:AP}]{\includegraphics[height=0.18\textheight,width=.44\textwidth]{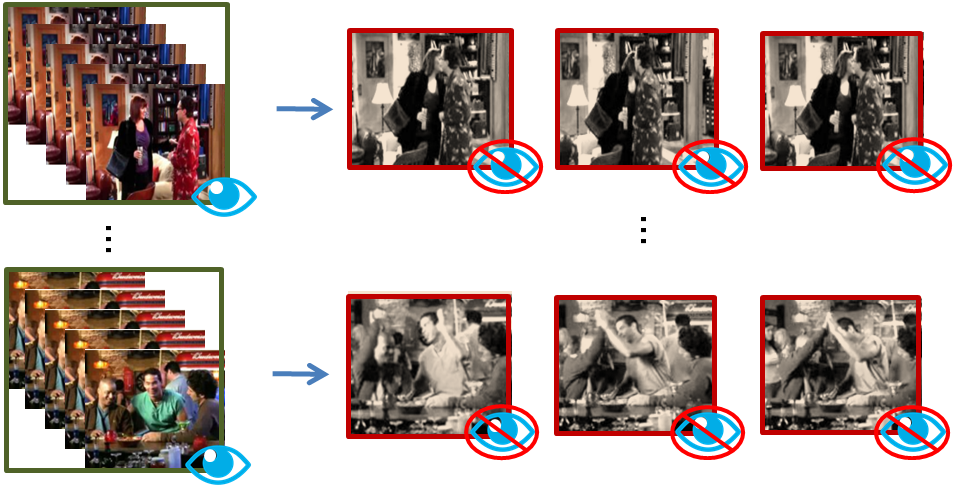}}\qquad\quad
	\subfigure[Intention prediction \label{s:IfM} ]{\includegraphics[height=0.18\textheight,width=.44\textwidth]{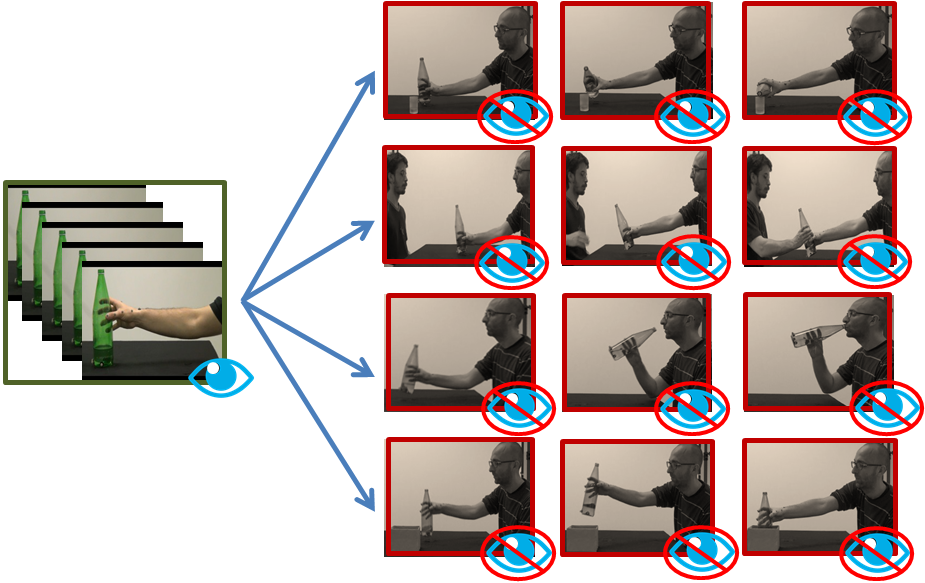}}
	\caption{Four different paradigms. \ref{s:AR} Action/activity recognition: each sequence is fully observed to infer the class label (``running'' for the top sequence up to ``high-five'' for the bottom). \ref{s:EAR} Early activity recognition: only a few initial frames per sequence is observed and the goal is an early classification from these incomplete observations. \ref{s:AP} Action prediction: future actions are predicted analysing all past events which, in general, can be very different across different classes. Thus, in the top sequence a standing up activity leads to predict a ``kissing'', while, in the bottom, a conversation between a group of friends anticipates a ``high-five''. \ref{s:IfM} Intention prediction: a novel paradigm where unobserved future action are anticipated from the same class of motor act, all extremely similar in appearance, no matter what different ending will occur.} 
	\label{fig:ciao}
\end{figure*}


In Fig. \ref{fig:ciao}, we show the aforementioned paradigms for action/activity analysis, and our new introduced problem of intent prediction. The core aspect and novelty of our problem stands from the fact that, in this case, intents cannot easily be predicted using discriminant previous information extracted from a certain anticipative data stream since, unlike the other paradigms, such data displays 
the same class of motor act which can be performed with different intents. 
Nevertheless, in general, the prediction of intents still remains a manageable, yet rather complex problem, as we will show in the following. 

The prediction of intents takes (early) activity recognition 
to an extreme level since, while it is relatively easy to recognize different actions from different onsets, the same task is not obvious when onsets are pretty similar, like the motor act of grasping a bottle finalized either to drinking water from it or to pour water into a nearby glass, for instance. 

This fact may look strange since the operation of grasping an object is apparently the same motor act whatever the action one wants to do next, but recent studies in psychology and neuroscience \cite{Asuini:2014,Asuini:2015,Becchio_ScientReport} have actually shown that the sole kinematics of the motor act (grasping) contains information about how to disambiguate the actual future intent (e.g., drinking or pouring). This challenge was never tackled before in the computer vision and multimedia communities, and this work aims at investigating this problem from a computational perspective, also providing a quantitative assessment by analyzing multi-modal video and motion capture 3D data.
Providing insights on this new task is of high importance since transferring the capability of predicting actual intents 
to intelligent systems will lead to a wide range of meaningful applications. For instance, in video surveillance and security applications, it could be extremely valuable to guess if a person inserting an hand in its pocket may retrieve the wallet or a gun.
Furthermore, in autonomous driving cars, an on-board vehicle vision system could predict a possible accident from the surrounding movements (of people and objects around), eventually preventing or avoiding it. Moreover, such capability is of paramount significance in human-robot interaction.

Previous attempts in classifying future or unfinished actions utilize current developing motion patterns which are specific of the subsequent actions, since they contain some cues that undoubtedly help the recognition. 
For instance, if the goal is understanding whether two people are going to shake their hands or to give a high-five, by just looking at the first part of their interaction, a low wrist height can be an evidence of a handshaking \cite{Torralba:15,Savarese14}. 
Further, another important aspect of the entire activity recognition problem is that the current techniques are mainly exploiting the scene context to support the classification (\cite{Cao:CVPR13,FZ:CVPR14,Amit:ACCV14,Xie:2013,Walker,Pei:2011,Amit:ACCV14,Kilner:2011,vanElk:2014}, and \cite{Bubetal:2013}), i.e., the objects present in the scene and the knowledge about the possible actions associated to them are cues that can be utilized to help in making a correct inference of the ongoing action to be recognized. This information is necessary but may be insufficient to solve the issue or, worse, the context might not always be available or easily recognizable, being also misleading when the scene is too noisy or cluttered \cite{Stapel:12,ZeB:15}. In any case, an important source of information to disambiguate intents is provided by the kinematics of the movement, and this work just wants to tackle this specific problem from a computational perspective regardless of the context and by analyzing multi-modal data.

To sum up, in this paper, we propose the new problem of human intents' prediction, named {\bf Intention from Motion} (IfromM), in a context-free setting, where kinematics is the only available cue analyzed. Two challenging aspects differentiate this work from the current literature. 

$\bullet$ Grounding from the assumption that the same class of motor acts can be performed with different intents \cite{Becchio_ScientReport,Asuini:2014,Asuini:2015}, we want to analyse the movement onset of an apparently unrelated action (actually embedding the intent from the very beginning), the same for all intents, capturing those subtle motion patterns which are discriminant of the future action. 

$\bullet$ Unlike the main existing literature, we want to avoid the exploitation of any cue derived by the context, solely focusing on the kinematics of the movement. 

To this end, we designed an ad-hoc experiment to introduce a new dataset aimed at investigating the feasibility of inferring intent from motion. In this experiment, subjects were asked to grasp a bottle, in order to either 1) pour some water into a glass, 2) pass the bottle to a co-experimenter, 3) drink from it, or 4) place the bottle into a box. The dataset is composed of both 3D trajectories of 20 motion capture (VICON) 3D markers outfitted over the hand of the participants, and optical video sequences (lasting about one seconds) with an occlusive camera view, in which only the arm and the bottle are visible. Data are acquired from the moment when the hand starts from a stable fixed position up to the reaching of the object, and 3D marker trajectories and video sequences are exactly trimmed at the instant when the hand grasps the bottle, removing the following part. We posit the problem as a multi-class classification task where the goal is to classify the intents associated with the observed grasping-a-bottle movement, \ie to predict the subject's intent.

Unlike previous work on activity recognition and prediction \cite{Savarese14,Ryoo,Torralba:15,Hoai,Walker}, the IfromM dataset is explicitly designed for predicting intents in a controlled setup which allowed us to accurately assess actual prediction capacity of computational methods. To conclude, the main contributions of this work are summarised in the following.


\begin{enumerate} 
	\item We introduce the new problem of Intention from Motion: from the same observable ``neutral'' motor act - used in both training and test phases - we try to classify the underlying intent using solely motion information, without exploiting any contextual cue.	
	\item  We design a principled experimental setup by defining four intents (Pouring, Passing, Drinking, Placing) performed by independent neutral subjects, which are all forerun from the very similar initial movement of grasping-the-bottle, while avoiding bias which can affect the subsequent performance analysis. Acquisitions is performed in a multi-modal way, providing both marker 3D trajectories and 2D RGB videos.  	
	\item While finding that the 2D video-based representation for action recognition/prediction are less effective with respect to the 3D one, we also discover that the two sources of information are actually complementary. This allows us to exploit fusion methods and achieve a reliable classification performance. Ultimately, this certifies the superiority of our multimodal approach to reliably predict intents in an exclusively kinematic (thus context-free) manner.	
\end{enumerate}

The rest of the paper is structured as follows. In Section \ref{sez:RL}, we report some related previous works. Section \ref{sez:dataset} introduces our experimental setting. The classification results obtained on the 3D markers and on the 2D videos are discussed in Section \ref{sez:unim}. Section \ref{ss:fusion} presents our results obtained by our multimodal approach and, finally, Section \ref{sez:con} draws the conclusions and sketches future extensions.

\section{Related Work}\label{sez:RL}

In this Section, we briefly report the most relevant works from the existing literature, which deals with both early activity recognition and action prediction. 

Ryoo \cite{Ryoo} devise a system to infer the ongoing activity by only analysing its \emph{onset}, \ie its beginning. This is done with a dynamic programming method to match an extension of classical bag-of-features representation which allows to capture the temporal correlation of descriptors. Hoai and De la Torre \cite{Hoai} design a max-margin event detectors to address the problem of the early recognition of a specific human emotion after it starts but before it ends. Yu \ea \cite{Yu:ACM12} propose a local approach to categorize actions from their beginning. The temporal-dependencies between different spatial location are implemented into a probabilistic graphical model fed by histogram features. Cao \ea  \cite{Cao:CVPR13} split a complete action into temporal segments which are further represented by means of sparse coding, so that actions are recognizable from incomplete data. Ryoo \ea  \cite{Ryoo:HRI15} tackle early activity recognition from egocentric videos: the task is detecting the so-called \emph{onset signature}, a bunch of kinematic evidence which has strong predictive properties about the last part of the observed action.  Some works have attempted to investigate how much of the whole action is necessary to perform a classification: Davis and Tyagi \cite{Davis:imavis06} adopt a generative probabilistic framework to deal with the uncertainty due to limited amount of data, while Schindler and Van Gool \cite{Schindler:CVPR08} try to answer the aforementioned question using a similarity measure between the statical and the motion information extracted from videos. Soran \ea \cite {Soran:ICCV15} devise a notification system for daily activities where, for instance, the detection of an ongoing milk boiling alerts the human user. Xu \ea \cite{autocompletion} subdivide the beginning of a video into a bunch of snippets, and the final ending is predictable through a ranking model which simulates Internet query auto-completion. The early recognition is also tackled by Soomro \ea \cite{Soomro} by combining a conditional random field data representation with SVM for the prediction; Ma \ea \cite{disney} approach the same problem by combining LSTM and CNN architectures. Kong and Fu \cite{KongFuPAMI16} cast SVM as an action prediction machine by building a composite kernel on top of a dense extraction of spatio-temporal features. 
Li \ea \cite{Li:ECCV12} use a random tree to model all the kinematics up to a certain instant, thus constraining the prediction of the most likely action (\textit{e.g.}, predicting ``grab an object'' if ``reach an object'' is detected).  
Huang \ea \cite{Huang:ECCV14} face activity forecasting for human interactions: the acts of an agent induce a cost topology over the space of reactive poses where the response of the co-agent can be retrieved. Lan and Savarese \cite{Savarese14} develop the so-called \emph{hierarchical movemes} to model human actions at multiple levels of granularities. 
Finally, deep learning approaches have been proposed by Vondrick and Torralba \cite{Torralba:15}, Jain \ea \cite{Jain16} and Fermuller et al. \cite{Fermuller}, using convolutional, recurrent or LSTM networks, respectively. However, the context is exploited by \ea \cite{Jain16} and the future action is either used during training \cite{Torralba:15} or to achieve a reliable performance \cite{Fermuller}.

One common aspect of both (early) activity recognition and action prediction is that contextual information is frequently used to perform the classification. Indeed, once the objects present in a scene are detected, the object-object or object-person relationship can be modelled by several probabilistic architectures (\eg, graphical models \cite{Amit:ACCV14,FZ:CVPR14,Li:PAMI14} or topic models \cite{Koppula,Pei:2011}). Among the works which directly model the context inside the algorithms, some of them deal with the prediction of future trajectories of moving objects (vehicles or pedestrian) \cite{Kitani,Walker,YamaguchiCVPR11,Xie:2013} by estimating the spatial areas over which such objects will most likely pass with respect to those which are excluded by this passage (\eg, car circulations over sidewalks \cite{Walker}).

In this paper, unlike all the aforementioned works, we are not classifying actions from their very first beginning (e.g., \cite{Ryoo,Hoai}) nore exploiting the future action to boost the performance \cite{Torralba:15,Fermuller}. Differently, we aim at predicting \emph{intent from motion}, a brand new challenge in action prediction consisting in predicting intents which finalize the same class of motor act. We distill from it the discriminative motion patterns characterizing the specific intent, while fully neglecting any contextual information (as opposed to \cite{Amit:ACCV14,Koppula,Walker}).

\section{Experimental setup}\label{sez:dataset}

Seventeen neutral volunteers were seated beside a $110 \times 100$ cm table resting on it elbow, wrist and hand inside a tape-marked starting point. A glass bottle was positioned on the table at a distance of about $46$ cm and participants were asked to grasp it in order to perform one of the following 4 different intents.
\begin{enumerate}
	\item {\textit{Pouring}} some water into a small glass (diameter $5$ cm; height $8.5$ cm) positioned on the left side of the bottle, at $25$ cm from it.
	\item{\textit{Passing}} the bottle to a co-experimenter seating opposite the table.
	\item{\textit{Drinking}} some water from the bottle.
	\item {\textit{Placing}} the bottle in a cardboard $17 \times 17 \times 12.5$ box positioned on the same table, $25$ cm distant.
\end{enumerate}
After a preliminary session, in which participants are familiarized with the execution, each subject performed $20$ trials per intent. The experimenter visually monitored each trial to ensure exact compliance of these requirements. During the recording stage, we completely removed trials judged imprecise and the final dataset includes $1098$ trial ($253$ for pouring, $262$ for passing, $300$ for drinking and $283$ for placing) and, for each of them, both marker 3D trajectories and 2D video clips have been collected.

\begin{figure}[t!]
	\centering
	\includegraphics[width=\columnwidth,keepaspectratio]{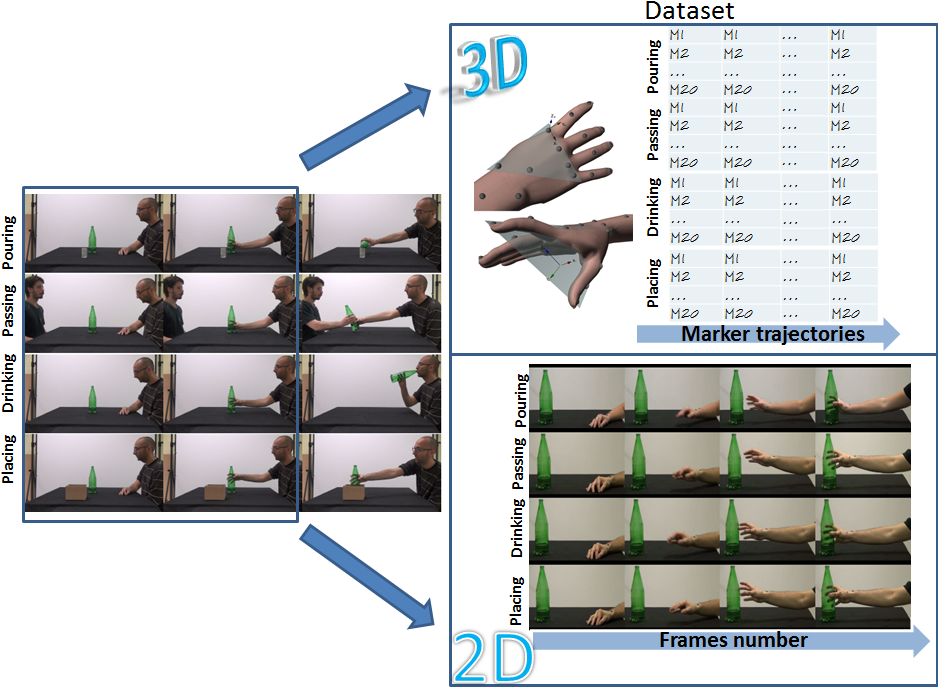}%
	\caption{Experimental setup. On the left we have the entire visible pouring, passing, drinking and placing development. On the top right, we have 3D VICON data acquisition, on bottom right, video sequences in which camera shoots only the arm and the bottle. In both cases, the acquisition stop at the grasping moment.}	
\end{figure}

\noindent
\underline{3D kinematic data.} Near-infrared $100$ Hz VICON system was used to track the hand kinematics. Nine cameras were placed in the experimental room and each participant's right hand was outfitted with $20$ lightweight retro-reflective hemispheric markers. After data collection, each trial was individually inspected for correct marker identification and then run through a low-pass Butterworth filter with a 6 Hz cutoff. Globally, each trial is represented with a set of 3D points describing the trajectory covered by every single marker during execution phase. The $x,y,z$ marker coordinates only consider the reach-to-grasp phase, where the following movement is totally discarded. Indeed, the acquisition of each trial is automatically ruled by a thresholding of the wrist velocity $v(t)$ at time $t,$ acquired by the corresponding marker. Being $\varepsilon = 20$ mm/s, at the first instant $t_0$ when $v(t_0) > \varepsilon,$ the acquisition starts and it is stopped at time $t_f,$ when the wrist velocity $v(t_f) < \varepsilon.$ 

\noindent
\underline{2D video sequences.} Movements were also filmed from a lateral viewpoint using a fixed digital video camera (Sony Handycam 3-D) placed at about $120$ cm from hand start position. The view angle is directed perpendicularly to the agent's midline, in order to ensure that the hand and the bottle were fully visible from the beginning up to the end of the movement. It is worth noting that the video camera was positioned in a way that neither the box (placing), nor the glass (pouring), nor the co-experimenter (passing) were filmed. Adobe Premiere Pro CS6 was used to edit the video in .mp4 format with disabled audio, 25 fps and  $1280 \times 800$ pixel resolution. In order to format video sequences in an identical way to 3D data, each video clip was cut off at the exact moment when the bottle is grasped, discarding everything happening afterwards. To better understanding how demanding the task is, note that the actual acquired video sequences encoding the grasping last for about one fourth of the future action we want to predict. Consequently all the sequences result about 30 frames long.

Before proceeding, let us conclude with two additional remarks.

$\bullet$ {\bf Challenging one-subject-out testing strategy}. In all the experiments reported in this paper, either dealing with 3D or 2D data, we consider all the possible pairwise comparisons between intents and the all-class one. We select \textbf{\textit{one-subject-out}} testing procedure, that is, we compute seventeen accuracies, training our system on all the subjects except the one we are testing, then we averaged all the accuracies to get the final classification results.

$\bullet$ {\bf Relationships with existing experimental setups}. If compared with the existing literature, the controlled experimental conditions of our setup seems a limitation. For instance, MPII-CAD \cite{MPII} and Salad 50 \cite{salad} cover more articulated (cooking) actions, while UCF-101 \cite{UCF101} and HDMB51 \cite{HDMB51} collect YouTube videos, thus guaranteeing a broad variability of backgrounds and context. Conversely, we deliberately designed our case study in order to properly answer the question if the kinematics of the same ongoing action is enough informative to discover the intent which caused the following action. 
Indeed, the uncontrolled and real-world scenarios of the YouTube videos (such as in UCF-101 and HDMB51) may accidentally enrich the context with some cues which actually facilitate the prediction. Moreover, different future actions frequently begin with a quite different onset, \eg, two persons \emph{approach} each other before a ``kissing'' action occurs, or people \emph{rise their hands} before a ``high-five'' action is carried out \cite{Savarese14}. Additionally, in MPII-CAD and Salad 50 for instance, the prediction is facilitated by the detection of \emph{which} objects (out of many others) is grasped (\eg, a knife to predict ``cutting''). Conversely, we want to predict \emph{why} the same object (bottle) is grasped by inspecting \emph{how} the latter action is accomplished under an exclusive kinematic point of view. 

\section{Unimodal analyses}\label{sez:unim}

\begin{table*}[t!]
	\centering
	\caption{ 3D results. When SVM is used, we fixed its cost parameter $C=10$. We performed a nearest neighbors classification with $K=5$. For H-COV, we used the default choice parameter $L=3$ with overlap (see \cite{egizi}). We selected the ker-COV \cite{kerCOV} parameter after cross-validation.}
	\label{tab:3D}

	\begin{tabular}{|c|c|c|c|c|c|c|c|}
		{\bf 3D results} & \multicolumn{3}{|c|}{Kinematic Features} & \multicolumn{2}{c|}{DTW} & \multicolumn{2}{c|}{Covariance-based} \\
		(\%) & $F_{\rm local}$   & $F_{\rm global}$ & $F_K$ & $K$-nn & $\mathcal{L}$ + SVM & H-COV & ker-COV   \\ \hline\hline
		Pouring vs. Placing & 79.70 & 86.10 & 84.32 &   87.86 & 83.28 & 90.23 & {\bf 91.87}	\\
		Pouring vs. Drinking & 72.15 & 70.36 & 76.48 &  67.06 & 83.59 & 91.43 & {\bf 91.58} \\
		Pouring vs. Passing & 76.55 & 67.39 & {\bf 82.81} &  66.49 & 81.98 & 80.43 & 81.69 \\
		Passing vs. Drinking & 63.10 & 68.05 & 70.75 & 54.29 & 82.53 & 87.30	& {\bf 87.64} \\
		Passing vs. Placing & 62.60 & 64.38 & 69.44 & 64.37 & {\bf 82.27} & 76.50 & 75.46  \\
		Drinking vs. Placing & 64.40 & 71.41 & 73.72 & 71.51 & 90.74 & 89.63	&  {\bf 91.24} \\\hline
		All-class    & 45.08 & 48.01 & 55.13 & 40.86 & 63.10 & 70.82 & {\bf 73.72}
	\end{tabular}
\end{table*}

In this Section we explain how we encode the 3D kinematics acquired with motion capture (Section \ref{sez:3D}) and the 2D RGB video frames (Section \ref{sez:2D}). Finally, we jointly discuss the results achieved by the two modalities in Section \ref{sez:unimdisc}.

\subsection{Analysis of marker 3D trajectories}\label{sez:3D}

Several techniques have been proposed for action recognition from 3D data: bag-of-points \cite{Li:CVPRw10}, eigen-joints \cite{Yang:CVPRw12}, Gauss-Markov process \cite{Chaudhry:CVPRw13}, actionlets \cite{Wang:CVPR12}, Lie algebra embedding \cite{Vemulapalli:CVPR14}, covariance descriptors \cite{egizi}, hidden Markov models  \cite{Lv:ECCV06}, subspace view-invariant metrics \cite{Sheikh:ICCV05} or occupancy patterns \cite{Wang:ECCV12}, to name a few.

Nevertheless, to the best of our knowledge, no attempt has been performed to address the problem of action prediction from 3D data. Thus, in this Section, we will analyse the markers trajectories in our dataset with kinematic features, Dynamic Time Warping and covariance-based representations. 

\noindent
\underline{Kinematic Features.} Following \cite{Carpinella:2011}, we computed \emph{wrist velocity}, the module of the velocity of the wrist marker, \emph{wrist height}, the $z$-component of the wrist marker, \emph{wrist horizontal trajectory} defined as the $x$-component of the wrist marker and \emph{grip aperture}, \textit{\ie} the distance thumb-index tips markers. Such features were referred to the motion capture reference system, $F_{\rm global}$ \cite{Carpinella:2011}. A better characterization of the dynamics can be provided using a local reference system centered on the hand, $F_{\rm local}$ \cite{Asuini:2015}. In this way, we computed relative $x,y,z$ coordinates of thumb, index, thumb-index plane and the radius-phalanx. These variables provide the information about either the adduction/abduction movement of the thumb and index fingers or the  rotation of the hand dorsum. Thus, they ensure robustness towards finger flexion/extension or wrist rotation that can vary significantly from one trial to another \cite{Asuini:2015}. The 4 features from $F_{\rm global}$ and the 12 from $F_{\rm local}$ gives a total amount of 16 kinematic features that we can concatenate in $F_{\rm k}$. Acquisition time $[t_0,t_f]$ (see Section \ref{sez:dataset}) is scaled into $[0,1]$ and data are sub-sampled with step 0.01. Consequently, for each of our kinematic features, we have $100$ equispaced values describing the evolution of such features during the reach-to-grasp movement: globally, $F_{\rm local}$, $F_{\rm global}$ and $F_K$ shapes as a 1200, 400 and 1600-dimensional descriptor, respectively, which fed a linear support vector machine (SVM).

\noindent \underline{Dynamic Time Warping (DTW).} We used DTW to construct a similarity measure $\Delta$ between multivariate time-series, exploiting the notion of alignment through warping paths (see \cite{DTW}). Thus, after computing $\Delta$ for all pairs of motion sequences from our dataset, we got the 1098 $\times$ 1098 distance matrix which was both directly used as metric for $K$-nearest neighbours ($K$-nn) classification and converted into a kernel by means of the graph Laplacian operator $\mathcal{L}$ to feed SVM classification \cite{negDTW}.

\noindent \underline{Covariance-based paradigms.} We inspected the sampling covariance estimator - briefly, covariance - in predicting human intents from motion since in the field of action recognition from motion capture (MoCap) systems, many works were actually based on such kind of representation. For instance, \cite{egizi} proposed a hierarchical model composed by a L-layered temporal pyramid of covariance descriptors (H-COV). Also, in the recent work \cite{kerCOV}, the new state-of-the-art in action recognition from MoCap data as obtained by a rigorous kernelization of the covariance operator (ker-COV) in order to model, general, non-linear, temporal correlations of marker coordinates. In both cases, we vectorized either the temporal pyramid \cite{egizi} or the kernelization \cite{kerCOV}, feeding it into a linear SVM.

\begin{table*}[t!]
	\caption{ DT features for SVM classification ($C=10$). For \emph{BoF}, we computed an exponential $\chi^2$ kernel, while, for \emph{FV} and \emph{VLAD}, a linear kernel was adopted.}
	\begin{tabular}{|c|rl|rl|rl|rl|rl|rl|}
		\centering
		{\bf DT results} & \multicolumn{2}{|c|}{HOG (\%)}   & \multicolumn{2}{|c|}{HOF (\%)} & \multicolumn{2}{|c|}{TSD (\%)} & \multicolumn{2}{|c|}{MBHx (\%)}  & \multicolumn{2}{|c|}{MBHy (\%)}  & \multicolumn{2}{|c|}{HOT (\%)} \\ \hline \hline
		Pouring  & \emph{BoF}   &  85.28 & \emph{BoF}     & {\bf 85.75} & \emph{BoF} & 83.23 & BoF   & 85.56 & \emph{BoF} & 84.64 & \emph{BoF} & 83.64 \\
		vs. & \emph{FV}     &   87.12 & \emph{FV}    &  {\bf 87.59} & \emph{FV}    & 78.84 & \emph{FV}    & 85.96 & \emph{FV}    & 83.06 & \emph{FV}    & 78.64 \\ 
		Placing & \emph{VLAD} & 86.71 & \emph{VLAD} & 86.18    & \emph{VLAD} & 81.60 & \emph{VLAD} & {\bf 87.65} & \emph{VLAD} & 85.70 & \emph{VLAD} & 76.93 \\ \hline
		Pouring &  \emph{BoF} & 70.63 &  \emph{BoF} &  {\bf 77.21} & \emph{BoF} & 72.66 & \emph{BoF} & 70.78 & \emph{BoF} & 74.15 & \emph{BoF} & 64.81 \\ 
		vs. & \emph{FV}     & 75.48  & \emph{FV}    &  {\bf 81.03}  & \emph{FV}    & 73.39 & \emph{FV} & 76.73 & \emph{FV} & 75.46 & \emph{FV} & 62.40 \\ 
		Drinking & \emph{VLAD} & 77.33  & \emph{VLAD} &  {\bf 81.48}  & \emph{VLAD} & 77.92 & \emph{VLAD} & 74.61 & \emph{VLAD} & 76.22 & \emph{VLAD} & 60.23 \\ \hline
		Pouring & \emph{BoF} & 71.77 & \emph{BoF} & 67.41 & \emph{BoF} & 72.17 & \emph{BoF} & {\bf 72.55} & \emph{BoF} & 68.33 & \emph{BoF} & 72.22 \\
		vs. & \emph{FV} & 75.90 & \emph{FV}  &  {\bf 79.75}  & \emph{FV} & 67.16 & \emph{FV} & 75.15 & \emph{FV} & 70.17 & \emph{FV} & 65.58 \\
		Passing & \emph{VLAD} & {\bf 77.48} & \emph{VLAD} &  74.44  & \emph{VLAD} & 73.58 & \emph{VLAD} & 76.01 & \emph{VLAD} & 68.48 & \emph{VLAD} & 63.14 \\ \hline
		Passing & \emph{BoF} & 66.67 & \emph{BoF} & 65.49 & \emph{BoF} & 65.34 & \emph{BoF} & {\bf 71.21} & \emph{BoF} & 64.65 & \emph{BoF} & 69.10 \\
		vs. & \emph{FV} & 66.88 & \emph{FV} &  {\bf 73.22}  & \emph{FV} & 61.44 & \emph{FV} & 68.45 & \emph{FV} & 68.02 & \emph{FV} & 64.69 \\
		Drinking & \emph{VLAD} & 70.06 & \emph{VLAD} & {\bf 71.53}   & \emph{VLAD} & 67.68 & \emph{VLAD} & 69.78 & \emph{VLAD} & 66.25 & \emph{VLAD} & 61.02 \\ \hline
		Passing & \emph{BoF} & {\bf 66.99} & \emph{BoF} & 66.34 & \emph{BoF} & 58.28 & \emph{BoF} & 65.25 & \emph{BoF} & 59.57 & \emph{BoF} & 66.64 \\
		vs. & \emph{FV} & 65.55 & \emph{FV} & {\bf 76.84}   & \emph{FV} & 56.00 & \emph{FV} & 66.86 & \emph{FV} & 60.02 & \emph{FV} & 63.75 \\
		Placing & \emph{VLAD} & 65.83 & \emph{VLAD} &  {\bf 75.15}  & \emph{VLAD} & 61.62 & \emph{VLAD} & 67.85 & \emph{VLAD} & 63.80 & \emph{VLAD} & 62.63 \\ \hline
		Drinking& \emph{BoF} & 70.66 & \emph{BoF} & {\bf 75.60} & \emph{BoF} & 68.77 & \emph{BoF} & 73.19 & \emph{BoF} & 71.60 & \emph{BoF} & 70.44 \\
		vs. & \emph{FV} & 73.04 & \emph{FV} & {\bf 78.41} & \emph{FV} & 73.99 & \emph{FV} & 74.04 & \emph{FV} & 73.04 & \emph{FV} & 65.05 \\
		Placing & \emph{VLAD} & 72.55 & \emph{VLAD} & {\bf 79.23} & \emph{VLAD} & 72.24 & \emph{VLAD} & 75.35 & \emph{VLAD} & 73.27 & \emph{VLAD} & 63.84 \\ \hline
		\multirow{3}{*}{All-class} & \emph{BoF} & {\bf 48.16} & \emph{BoF} & 48.05 & \emph{BoF} & 45.01 & \emph{BoF} & 47.71 & \emph{BoF} & 45.80 & \emph{BoF} & 46.33 \\
		& \emph{FV} & 50.02 & \emph{FV} &  {\bf 56.97}  & \emph{FV} & 46.00 & \emph{FV} & 50.35 & \emph{FV} & 47.23 & \emph{FV} & 41.12 \\
		& \emph{VLAD} & 51.88 & \emph{VLAD} & {\bf 58.23} & \emph{VLAD} & 51.63 & \emph{VLAD} & 53.30 & \emph{VLAD} & 48.63 &\emph{VLAD}  & 38.62      
	\end{tabular}
	\label{tab:2D}
\end{table*}

\subsection{Analysis of 2D video sequences }\label{sez:2D}

Far from providing a comprehensive review of the whole action recognition/prediction literature on video data, in this Section, we will benchmark the best hand-crafted descriptors (dense trajectories \cite{DTjournal}) as well as 3D Convolutional Neural Network (CNN) features for video representation, and frame-based deep encodings. 
Moreover, we will show that many currently available frameworks in early activity recognition and action prediction are not suitable for our test-bed problem.

\noindent \underline{Dense trajectories (DT).} Being part of the class of approaches named in \cite{arxiv} as local, DT \cite{DTjournal} track in time a set of spatio-temporal interest points (IPs) from an input video, using a dense optical flow field. For each IP, its trajectory is surrounded by a warped volume from which we computed classical histogram features: Histograms of Oriented Gradients (HOG) \cite{HOG}, Histograms of Optical Flow (HOF) \cite{HOF}, Motion Boundary Histograms \cite{MBH} in both $x$ and $y$ directions (MBHx and MBHy), trajectory shape descriptor (TSD) \cite{DTjournal} and Histograms of Oriented Trackelets (HOT) \cite{HOT}. We used the publicly available DT code\footnote{\url{http://lear.inrialpes.fr/software/}}, adopting the default parameters choice except to the trajectory length which was set to $5$ to better deal with our extremely short footages. 

In order to combine the dense histogram features into a unique video descriptor, we either applied $\ell^1$ normalized bag-of-features histograms (\emph{BoF}) \cite{Boiman08}, square-root normalized Fisher Vector (\emph{FV}) \cite{Perronin2007}, or Vectors of Locally Aggregated Descriptors (\emph{VLAD}) \cite{Jegou2010}. For \emph{BoF} and \emph{VLAD}, we used a dictionary of 1000 visual words; for \emph{FV} we employed a Gaussian mixture model with 256 components (as in \cite{Wang:improved}).

\begin{table}[h!]
	\centering
	\caption{ Accuracies of pre-trained C3D and fine-tuned AlexNet architectures used as feature extractors for a subsequent SVM classification ($C=10$). When using \emph{BoF} and \emph{VLAD}, the size of the dictionary was fixed to 1000 and 50 respectively.}
	\label{tab:CNN}
	\begin{tabular}{|c|c|c|c|c|c|c|}
		\multirow{2}{*}{\bf CNN results} & \multicolumn{2}{|c|}{C3D} & \multicolumn{3}{c|}{AlexNet} \\ 
		& I  &  OF   &    I  & \multicolumn{2}{|c|} {OF}  \\
		(\%) & & & \emph{BoF} & \emph{BoF} & {\em VLAD} \\ \hline\hline
		{\scriptsize Pouring vs. Placing} & 83.15 & 87.34  & 74.01 & {\bf 94.58}  & 94.18	\\
		{\scriptsize Pouring vs. Drinking} & 68.20 & 68.93 & 62.44 & 74.93 & {\bf 77.95} \\
		{\scriptsize Pouring vs. Passing} & 69.42 & 65.02 & 60.28 & {\bf 75.89} & 74.20 \\
		{\scriptsize Passing vs. Drinking} & 61.57 & 61.05 & 55.73 & 61.41 & {\bf 66.05} \\
		{\scriptsize Passing vs. Placing} & 66.84 & 78.10 & 62.09 & {\bf 96.23} & 94.68 \\
		{\scriptsize Drinking vs. Placing} & 68.30 & 76.67 & 58.69 & 95.87 & {\bf 96.18} \\\hline
		All-class    & 45.51 & 52.14 & 37.03 & 64.55 & {\bf 65.64}
	\end{tabular}
\end{table}

\noindent \underline{CNN features.} We applied the three-dimensional convolutional network architecture C3D proposed in \cite{C3D}. Thus, we divided each video sequence in three clips of 16 frames, where each of them is codified with \texttt{fc6} features. The video descriptor simply concatenates the three representations of the clips into a $3\times4096$ vector, finally used to train a linear SVM. As input clips, we have considered stacks of raw frames (I), also representing the optical flow (OF) magnitude computed between pairs of consecutive frames. 


Similarly to \cite{Ng:CVPR15,Zha:BMVC15} we exploited CNNs for a frame-wise representation, employing, as a first setup, the AlexNet architecture \cite{imagenet} once fine-tuned on our video frames. Precisely, we extracted \texttt{fc7} features from all single frames I and, consequently, we encoded each video with \emph{BoF} as in \cite{Xu:CVPR15}. In a second experiment, in order to better capture the kinematics of the graspings, we fed AlexNet with OF images after another preliminary fine-tuning to match the new type of data. Inspired by \cite{PCNN}, in this case, we computed OF images with three channels constituted by the horizontal, vertical component and the magnitude of the optical flow field, after a preliminary normalization in the range $[0, 255]$. To obtain the descriptor for each video, we either applied \emph{BoF} and \emph{VLAD} encoding upon the AlexNet-OF deep representation.

\subsection{Discussion}\label{sez:unimdisc}	In Table \ref{tab:3D}, we report the results obtained with all the 3D encodings considered. As expected, when we combine $F_{\rm local}$ and $F_{\rm global}$ in $F_K$ the performance generally improves. Globally, $K$-nn using DTW is worse than $F_K$ with the exception of Pouring vs. Placing. With respect to the $K$-nn approach, the graph Laplacian $\mathcal{L}$ allows to boost the DTW classification performance in almost all the binary/all-class comparisons. A further boost in accuracy is provided by H-COV \cite{egizi} and ker-COV \cite{kerCOV} with an all-class improvement of 7.72\% and 10.62\% respectively. 

In Table \ref{tab:2D} and \ref{tab:CNN}, we report the performance of hand-crafted and data-driven representations for the 2D data, respectively. With the exception of HOT representation \cite{HOT}, all hand-crafted histogram-based encodings are, generally speaking, equally performant (Table \ref{tab:2D}), being sometimes much more effective than spatio-temporal convolutions of \cite{C3D}: VLAD-HOF scores 58.23\% on the all class case, which is about 6\% better than space-time architecture of \cite{C3D} fed with OF images. Additionally, the usage of a fine-tuned AlexNet architecture is actually able to improve the previous score (in tandem with VLAD and OF, about +6\% over DT-HOF).

{\bf Snippet Analysis.} We present a temporal analysis of the reach-to-grasp motions to verify if, in the 3D/2D data, it is possible to find any peculiar instant which is richer in kinematic discriminants than others.

We performed a snippet analysis where the 3D marker trajectories and 2D video sequences were trimmed to cover the initial 20\%, 40\%, 60\%, 80\% or 100\% of the original grasping execution only. Please note that, on our dataset, the latter is extremely short (2 seconds on average): hence, the snippet analysis forces the classification to rely on a very limited information. For instance, at 20\%, for the shortest trial in our dataset, we have to use 16 markers acquisitions and 3 video frames only. As to monitor the impact of limiting the temporal domain on the 3D and 2D data separately, we considered the descriptors which obtained the best performance in the two baselines, respectively: ker-COV \cite{kerCOV} (Section \ref{sez:3D}) and AlexNet-OF-{\em VLAD} (Section \ref{sez:2D}). In this case, ker-COV representation only models temporal correlations among the initial 20\%, \dots, 100\% acquisitions of the markers and, similarly, the dictionary for {\em VLAD} encoding AlexNet-OF features is specific of the considered portion of the videos only. 

Table \ref{tab:perc} report the results of the snippet analysis for AlexNet-OF-{\em VLAD}. Therein, the best scores are obtained by considering high percentages (80\% and 100\%), but, anyway, we are able to capture discriminative information already from the beginning of the grasping: \textit{e.g.} $90.93 \%$ in Passing vs. Placing at 20 \%. Differently, in Table \ref{tab:Kerperc}, the snippet analysis with ker-COV does not remarkably exceed random chance level at 20\% and 40\%, with a great jump in performance at 80\% and 100\%. 

In spite of this, we can anyway find a common trend between the results of Tables \ref{tab:Kerperc} and \ref{tab:perc}. Namely, we registered a general growth in accuracy when the data percentages increase. Consequently, we can not find any portion of the reach-to-grasp execution that is completely useless for the prediction of intents. 

\begin{table}[t!]
	\centering
	\caption{ Results for the snippet analysis using ker-COV features.}
	\label{tab:Kerperc}
	\begin{tabular}{|l|c|c|c|c|c|c|c|c|c|c|c|}
		
		\centering
		{\bf \scriptsize 3D snippet analysis} & 20\%  & 40\% & 60\%  & 80\% & 100 \%\\\hline\hline
		{\scriptsize Pouring vs. Placing} & 51.48 & 71.40 & 76.89  & 87.55 & {\bf 91.87}\\
		{\scriptsize Pouring vs. Drinking}  & 47.85 & 61.01 & 64.45  & 72.30  &  {\bf 91.58}\\
		{\scriptsize Pouring vs. Passing} & 47.89  & 52.75  & 56.98  & 70.77  & {\bf 81.69}\\
		{\scriptsize Passing vs. Drinking} & 50.75  & 53.98  &  59.62  & 71.97  & {\bf 87.64}\\
		{\scriptsize Passing vs. Placing}  & 54.20 & 61.67  & 62.75 & 70.68  & {\bf 75.46}\\
		{\scriptsize Drinking vs. Placing}  & 54.48  & 60.00  & 64.93  & 67.84  & {\bf 91.24}\\\hline
		{\scriptsize All-class}   & 27.90  & 33.31  & 38.60  & 49.03  &  {\bf 73.72}
		
	\end{tabular}\vspace{.5 cm}
	
	\caption{ Results for the snippet analysis using AlexNet-OF-{\em VLAD} features.}
	\label{tab:perc}

	\centering
	\begin{tabular}{|l|c|c|c|c|c|c|c|c|c|c|c|}
		
		\centering
		
		{\bf \scriptsize 2D snippet analysis} & 20\%  & 40\% & 60\%  & 80\% & 100 \%\\\hline\hline
		{\scriptsize Pouring vs. Placing}  & 77.38 & 87.02  & 91.43  & 92.06  & {\bf94.18}\\
		{\scriptsize Pouring vs. Drinking}  & 61.31  & 67.49  & 71.52 & 73.31  &  {\bf 77.95}\\
		{\scriptsize Pouring vs. Passing} & 57.82  & 65.47 & 66.00 & 69.98  & {\bf 74.20}\\
		{\scriptsize Passing vs. Drinking}  & 61.25 & 62.85 &   65.25  & 65.62 & {\bf  66.05 }\\
		{\scriptsize Passing vs. Placing}  & 90.93 & 96.19  & 95.82  & {\bf 96.28 }  & 94.68\\
		{\scriptsize Drinking vs. Placing} & 89.42  & 95.00 & 95.42  & 95.18  & {\bf 96.18 }\\\hline
		{\scriptsize All-class}    & 49.70 & 57.79  & 57.91 & 62.27  &  {\bf 65.64}
	\end{tabular}
	
\end{table}

{\bf Evaluation of existing prediction pipelines.} It is worth noticing that several appproaches for 2D video-based action prediction approaches are not applicable to our experimental setup. Indeed, \cite{Li:ECCV12,Hoai,Savarese14} relies on a fine temporal tessellation of the onsets into short and discriminative snippets (movemes \cite{Savarese14}), while, in our case, we can rely on \emph{one} of such a snippet only. Since \cite{Torralba:15,Fermuller} manage to achieve a reliable performance when also observing the complete action, while we are not exploiting the pouring, passing, drinking or placing at all. Also, \cite{Walker,Koppula} are methods which massively rely on the context - we are context-free instead. Finally, we were only able to apply the dynamic bag-of-word histograms of \cite{Ryoo} to our case. Using this approach, the all-class classification accuracy is 45.12\%, which suffers a gap of $-$13.11\% and $-$20.52\% with respect to DT-HOF-\emph{VLAD} and AlexNet-OF-\emph{VLAD}, respectively. Thus, globally, despite all the aforementioned prediction pipelines are really effective in their experimental conditions, the same methods seem little generalizable to different settings (such as ours).  

Nevertheless, if comparing the scores of 2D approaches (Tables \ref{tab:2D}, \ref{tab:CNN} or existing action prediction pipelines) with the 3D alternatives (Table \ref{tab:3D}), we notice that the latter class is superior to the former one. A priori, there is no reason to postulate that a sparse and incomplete\footnote{For instance, elbow and shoulder dynamics are not modeled.} marker representation can be richer in discriminants than a global sequential information provided by the video data. Nevertheless, even hand-crafted 3D features ($F_K$) are able to score on pair to the deep space-time architecture of \cite{C3D}. This is a clear evidence of the effectiveness of our novel experimental apparatus for the multimodal prediction of human intents. 

Consequently, instead of the widely used video representation \cite{Ryoo,Li:ECCV12,Hoai,Savarese14,Torralba:15,Fermuller,Walker,Koppula}, an effective classification should exploit the 3D kinematics.

Actually, we can still query whether the two sources of informations can be combined in order to furthermore improve the model. This is what we investigate in Section \ref{ss:fusion}.


\section{Multimodal 3D + 2D Fusion}\label{ss:fusion}

A unique aspect of our proposed dataset refers to its multi-modal nature, namely providing both 3D markers trajectories and 2D video acquisitions of every reach-to-grasp onset. Thus, it is interesting to take advantage of such dual source of information to overcome the performance of methods which only leverage on one type of data only. If this happens, then the 3D and 2D data representations are clearly complementary.

To this aim, in this Section, we combine the techniques reported in Sections \ref{sez:3D} and \ref{sez:2D} by either merging the single descriptors (early fusion) or integrating kernel-based representation, each of them related to one 3D or 2D encoding at a time (late fusion). For a comprehensive analysis on fusion techniques, please refer to \cite{fusion_survey}.

\noindent \underline{Early fusion of feature vectors} -- Throughout our 3D and 2D baseline, several features have been envisaged: $F_K$, ker-COV \cite{kerCOV}, the six DT histogram descriptors and the deep representations extracted by either using C3D or AlexNet. In order to fuse all of them into a unique descriptor, we applied two techinques.

\noindent $(1)$ We concatenated all the aforementioned feature representations into a unique single vector and reduced its dimensionality from 579.786 to 160 components by means of PCA (38\% variance explained). \\ 
$(2)$ We applied the CMIM criterion \cite{brown} to capture the variability in the class label conditioned on the data, while also minimizing the redundancy with respect to previously selected component (see \cite{brown}). In our case, we used CMIM to select the 150 most discriminative feature components among all the different single representations from Sections \ref{sez:3D} and \ref{sez:2D}.

\noindent \underline{Late fusion of kernels} -- As the preliminary stage of our late fusion pipeline, we computed a kernel from each different data encoding separately: a Gaussian RBF kernel for each kinematic feature, the graph Laplacian for the DTW similarity matrix, a Gaussian $\chi^2$ kernel for AlexNet-I-\emph{BoF} and AlexNet-OF-\emph{BoF}. A linear kernel was used for ker-COV, C3D-I, C3D-OF and for the DT, AlexNet-OF features encoded with \emph{VLAD}. In order to train a SVM, the final kernel used is a linear combination of all the aforementioned ones, weighted according to the MSE and ACC criteria proposed in \cite{fusion_survey}. That is, each kernel is weighted according to the mean squared error (for MSE) and to the accuracy (for ACC) registered when using a SVM fed with that single kernel only. 

\begin{table}[t!]
	\centering
	\caption{Early fusion of feature descriptors and late fusion of kernels.}
	\label{tab:fus}
	\begin{tabular}{|c|c|c|c|c|c|}
		{\bf Fusion results} & BSD & \multicolumn{2}{c|}{\em Early Fusion} & \multicolumn{2}{c|}{\em Late Fusion} \\
		(\%) & & PCA & CMIM & MSE & ACC \\ \hline \hline
		{\scriptsize Pouring vs. Placing} & 94.18 & 85.80 & {\bf 95.92} & 88.84 & 95.70  \\
		{\scriptsize Pouring vs. Drinking} & 91.58 & 83.85 & 93.30 & 91.41 & {\bf 94.62}  \\
		{\scriptsize Pouring vs. Passing} & 84.47 & 79.88 & {\bf 90.47} & 85.85 & 90.04  \\
		{\scriptsize Passing vs. Drinking} & 87.75 & 84.89 & 87.21 & 82.57 & {\bf 90.30}  \\
		{\scriptsize Passing vs. Placing} & {\bf 96.23} & 70.23 & 93.49 & 82.33 & 91.12  \\
		{\scriptsize Drinking vs. Placing} & 96.18 & 82.63 & 93.68 & 90.97 & {\bf 96.87}  \\\hline
		{\scriptsize All-class} & 73.72 & 68.39 & 80.08 & 77.52 & {\bf 80.50}
	\end{tabular}\vspace{-.5 cm}
\end{table}

\subsection{Discussion}
The classification accuracies are reported in Table \ref{tab:fus}, where we also include the best single descriptor (BSD) among all the ones presented in Sections \ref{sez:3D} and \ref{sez:2D}. The early fusion is able to improve the performance of BSD on Pouring vs. Placing and Pouring vs. Passing. Similarly, with the exception of Passing vs. Placing, the late fusion improves all the remaining pairwise comparisons. Moving to the all-class case, the late fusion (ACC) and early fusion (CMIM) behaves in a similar manner, both overcoming the 80\% of classification accuracy. 

This is a strong experimental evidence that the 2D and 3D information are actually complementary in providing useful cues to the classification. Even if combining the 3D information with a less powerful data representation, the classification performance of the best either early of late fusion approach is about +7\% better than the best 3D descriptor (ker-COV \cite{kerCOV}).

In summary, despite the related challenge of the extreme similarity of grasping onsets (see the video material in attachment), the problem of a context-free intent prediction turns out to be reliably feasible by leveraging on multimodality.

\section{Conclusions}\label{sez:con}

In this paper, we propose the novel paradigm of intent prediction. That is, classifying different intents all starting with the same class of motor acts, without using contextual information. 

We have proposed a novel experimental apparatus where, in a neutral and uninformative context, a bottle is grasped in order to fulfill a pouring, passing, grasping or placing intent. Despite 1) the challenging experimental paradigms mine the applicability of existing prediction pipelines and 2) all the graspings seems very identical to each other at a first glance, we prove that the actual reaching-a-bottle pattern is specialized in function of the future intent to fulfill. No matter what 3D or 2D feature representation is used, random-chance in classification is overcome: this certifies that our novel problem is actually feasible.

If comparing 3D with 2D data representation, we find that marker trajectories as a kinematic encoder provide a more compact yet discriminative source of information with respect to 2D RGB videos. However, the two different modalities can be proficiently combined and, even if using classical fusion methods, the registered classification performance are extremely favorable (so that many binary comparisons are almost saturated).

Due to our one-subject-out experimental evaluation, we attest the feasibility of intent prediction systems which leverage on multimodality to proficiently generalize and reliably anticipate humans' intents in a pure kinematics-driven fashion.

Future directions refer to realize novel datasets to perform
intent prediction in social scenarios, dealing with more
composite actions.

\bibliographystyle{unsrt}
\bibliography{fonti}


\end{document}